\documentclass[10pt,twocolumn, letterpaper]{article} 

\newcommand{\final}{1} 
\newcommand{\forReview}{0} 

\usepackage{color}
\usepackage{ifthen}
\usepackage{float}
\usepackage{alltt}
\usepackage{newlfont} 
\usepackage{wrapfig}
\usepackage{booktabs}
\usepackage{multirow}
\usepackage{comment}
\usepackage{gensymb}
\usepackage{pifont}
\usepackage{xcolor}

\usepackage{comment}
\usepackage{amsmath}
\usepackage{amssymb}
\usepackage{amsthm}
\usepackage{iccv}
\usepackage{times}
\usepackage{epsfig}
\usepackage{graphicx}
\usepackage{amsmath}
\usepackage{amssymb}
\usepackage{amsthm}
\usepackage{subcaption}
\usepackage{microtype}
\usepackage{blindtext}

\definecolor{DeltaColor}{rgb}{0.039,0.73,0.71}
\definecolor{SetaColor}{rgb}{0.867, 0.0235, 0.376}
\definecolor{SigmaColor}{rgb}{0.98,0.45,0.0}
\definecolor{HaoColor}{rgb}{0.8,0,0}
\definecolor{AlphaColor}{rgb}{0,0,0.8}
\definecolor{BetaColor}{rgb}{0.8,0,0.8}
\definecolor{GammaColor}{rgb}{0.5,0,0.7}
\definecolor{EpsilonColor}{rgb}{0.353,0.725,0.906}
\definecolor{TauColor}{rgb}{0.423,0.235,0.192}
\newcommand{\shunsuke}[1]{{\color{AlphaColor} Shunsuke: #1 $\qed$}}
\newcommand{\ryota}[1]{{\color{SetaColor} Ryota: #1 $\qed$}}
\newcommand{\zeng}[1]{{\color{SigmaColor} Zeng: #1 $\qed$}}
\newcommand{\hao}[1]{{\color{GammaColor} Hao: #1 $\qed$}}
\newcommand{\angjoo}[1]{{\color{DeltaColor} Angjoo: #1}}

\newcommand{\warning}[1]{{\it\color{red} #1}}
\newcommand{\note}[1]{{\it\color{blue} #1}}
\newcommand{\nothing}[1]{}

\definecolor{AudioColor}{rgb}{0.56,0.34,0.62}

\definecolor{DeadlineColor}{rgb}{0.9,0.4,0} 
\newcommand{\deadline}[1]{{\bf\color{DeadlineColor} ETA: #1}}

\definecolor{figred}{rgb}{1,0,0}
\definecolor{figgreen}{rgb}{0,0.6,0}
\definecolor{figblue}{rgb}{0,0,1}
\definecolor{figpink}{rgb}{1,0.63,0.63}

\ifthenelse{\equal{\final}{1}}
{
\renewcommand{\l}[1]{}
\renewcommand{\shunsuke}[1]{}
\renewcommand{\zeng}[1]{}
\renewcommand{\ryota}[1]{}
\renewcommand{\hao}[1]{}
\renewcommand{\angjoo}[1]{}
\renewcommand{\warning}[1]{}
\renewcommand{\note}[1]{}
\renewcommand{\deadline}[1]{}
}
{}

\newcounter{pccount}
\floatstyle{plain}

\newcommand{\filename}[1]{\url{#1}}
\newcommand{\foldername}[1]{\url{#1}}

\usepackage{gensymb}

\ifthenelse{\equal{\forReview}{1}}
{
\usepackage[pagebackref=true,breaklinks=true,letterpaper=true,colorlinks,bookmarks=false]{hyperref}
\ificcvfinal\pagestyle{empty}\fi
}
{
\usepackage[breaklinks=true,colorlinks,bookmarks=false]{hyperref}
\iccvfinalcopy 

\newcommand*\samethanks[1][\value{footnote}]{\footnotemark[#1]}
\newcommand\blfootnote[1]{%
  \begingroup
  \renewcommand\thefootnote{}\footnote{#1}%
  \endgroup
}
\usepackage{cleveref}
\pagecolor{white}

\author{Shunsuke Saito\textsuperscript{1,2}\thanks{Joint first authors} \hspace{0.3in} Zeng Huang\textsuperscript{1,2}\samethanks \hspace{0.3in} Ryota Natsume\textsuperscript{3}\samethanks \\ 
Shigeo Morishima\textsuperscript{3} \hspace{0.3in}
Angjoo Kanazawa\textsuperscript{4} \hspace{0.3in} Hao Li\textsuperscript{1,2,5}
\vspace{5pt}
\\
\textsuperscript{1}{University of Southern California} \hspace{0.3in} \textsuperscript{2}{USC Institute for Creative Technologies} \\ \textsuperscript{3}{Waseda University} \hspace{0.3in} \textsuperscript{4}{University of California, Berkeley} \hspace{0.3in} \textsuperscript{5}{Pinscreen}
}
}

\graphicspath{
{figures}
{images}
}
\begin{document}

\title{PIFu: Pixel-Aligned Implicit Function for\\ High-Resolution Clothed Human Digitization\vspace{-0.4cm}}

\twocolumn[{%
  \renewcommand\twocolumn[1][]{#1}%
\maketitle
\begin{center}
  \newcommand{\teaserwidth}{\textwidth}
  \vspace{-0.15in}
  \centerline{
    \includegraphics[width=0.9\teaserwidth]{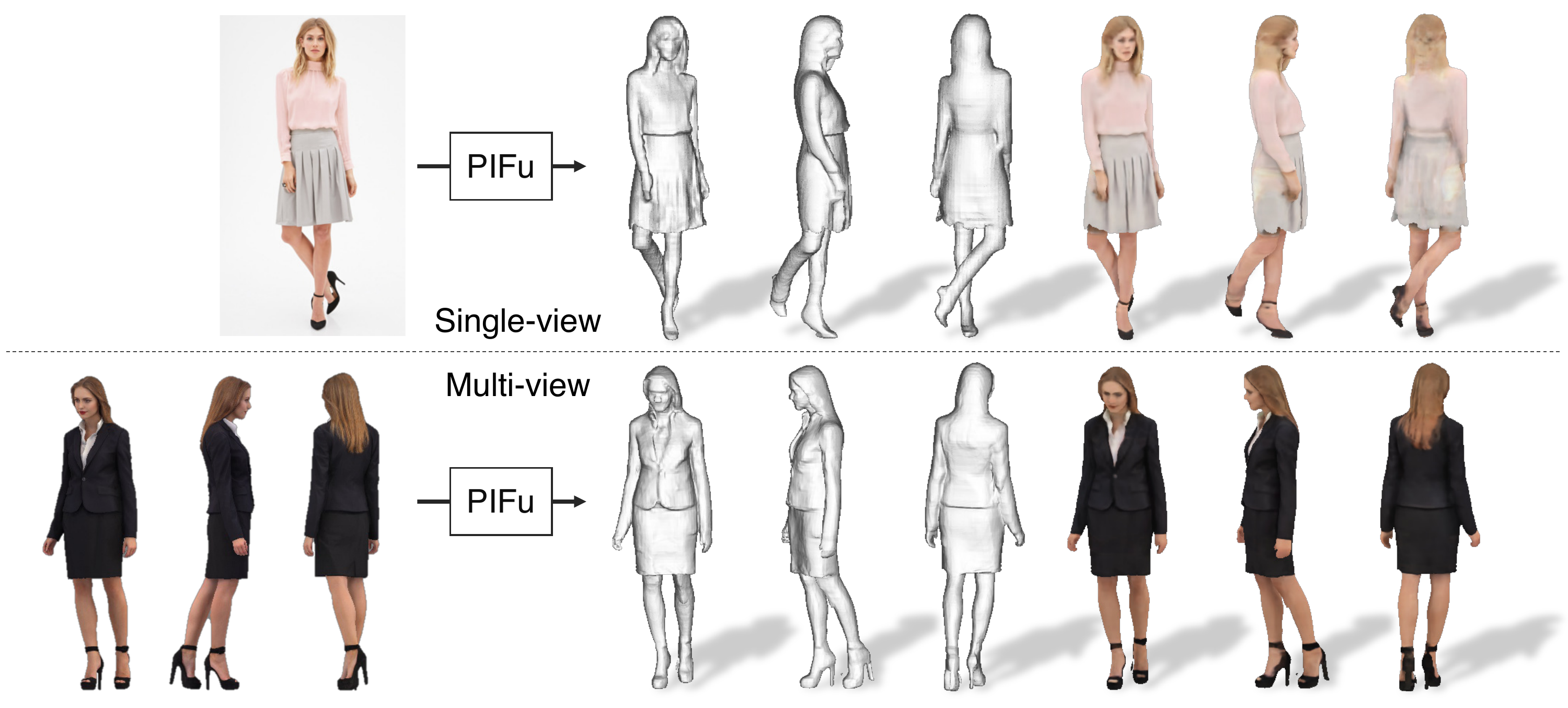}
    }
    \vspace{-10pt}
    \captionof{figure}{\textbf{Pixel-aligned Implicit function (PIFu): } 
    We present pixel-aligned implicit function (PIFu), which allows recovery of high-resolution 3D textured surfaces of clothed humans from a single input image (top row). 
    Our approach can digitize intricate variations in clothing, such as wrinkled skirts and high-heels, including complex hairstyles. The shape and textures can be fully recovered including largely unseen regions such as the back of the subject.
    PIFu can also be naturally extended to multi-view input images (bottom row).}
  \vspace{-0.05in}
  \label{fig:teaser}
 \end{center}%
    }]

\begin{abstract}
\vspace{-5pt}
\blfootnote{* - indicates equal contribution}
We introduce Pixel-aligned Implicit Function (PIFu), an implicit representation that locally \emph{aligns} pixels of 2D images with the global context of their corresponding 3D object. 
Using PIFu, we propose an end-to-end deep learning method for digitizing highly detailed clothed humans that can infer both 3D surface and texture from a single image, and optionally, multiple input images. Highly intricate shapes, such as hairstyles, clothing, as well as their variations and deformations can be digitized in a unified way. 
Compared to existing representations used for 3D deep learning, PIFu produces high-resolution surfaces including largely unseen regions such as the back of a person. In particular, it is memory efficient unlike the voxel representation, can handle arbitrary topology, and the resulting surface is spatially aligned with the input image.
Furthermore, while previous techniques are designed to process either a single image or multiple views, PIFu extends naturally to arbitrary number of views.
We demonstrate high-resolution and robust reconstructions on real world images from the DeepFashion dataset, which contains a variety of challenging clothing types. Our method achieves state-of-the-art performance on a public benchmark and outperforms the prior work for clothed human digitization from a single image. The project website can be found at \url{https://shunsukesaito.github.io/PIFu/}
\vspace{-10pt}

\end{abstract}

\section{Introduction}
\label{sec:intro}

In an era where immersive technologies and sensor-packed autonomous systems are becoming increasingly prevalent, our ability to create virtual 3D content at scale goes hand-in-hand with our ability to digitize and understand 3D objects in the wild. If digitizing an entire object in 3D would be as simple as taking a picture, there would be no need for sophisticated 3D scanning devices, multi-view stereo algorithms, or tedious capture procedures, where a sensor needs to be moved around. 

For certain domain-specific objects, such as faces, human bodies, or known man made objects, it is already possible to infer relatively accurate 3D surfaces from images with the help of parametric models, data-driven techniques, or deep neural networks. Recent 3D deep learning advances have shown that general shapes can be inferred from very few images and sometimes even a single input. However, the resulting resolutions and accuracy are typically limited, due to ineffective model representations, even for domain specific modeling tasks.

We propose a new Pixel-aligned Implicit Function (PIFu) representation for 3D deep learning for the challenging problem of textured surface inference of clothed 3D humans from a single or multiple input images. While most successful deep learning methods for 2D image processing (e.g., semantic segmentation \cite{fcn}, 2D joint detection \cite{tompson2014joint}, etc.) take advantage of ``fully-convolutional'' network architectures that preserve the spatial alignment between the image and the output, this is particularly challenging in the 3D domain. While voxel representations~\cite{varol18_bodynet} can be applied in a fully-convolutional manner, the memory intensive nature of the representation  inherently restrict its ability to produce fine-scale detailed surfaces. Inference techniques based on global representations~\cite{groueix2018atlasnet,hmrKanazawa17,alldieck2019learning} are more memory efficient, but cannot guarantee that details of input images are preserved. Similarly, methods based on implicit functions~\cite{chen2018implicit_decoder,park2019deepsdf,mescheder2018occupancy} rely on the global context of the image to infer the overall shape, which may not align with the input image accurately. 
On the other hand, PIFu aligns individual local features at the pixel level to the global context of the entire object in a fully convolutional manner, and does not require high memory usage, as in voxel-based representations. This is particularly relevant for the 3D reconstruction of clothed subjects, whose shape can be of arbitrary topology, highly deformable and highly detailed.
While \cite{Huang18ECCV} also utilize local features, due to the lack of 3D-aware feature fusion mechanism, their approach is unable to reason 3D shapes from a single-view. In this work we show that combination of local features and 3D-aware implicit surface representation makes a significant difference including highly detailed reconstruction even from a single view.

Specifically, we train an encoder to learn individual feature vectors for each 
pixel of an image that takes into account the global context relative to its position.
Given this per-pixel feature vector and a specified z-depth along the outgoing camera ray from this pixel, we learn an implicit function that can classify whether a 3D point corresponding to this z-depth is inside or outside the surface. In particular, our feature vector spatially aligns the global 3D surface shape to the pixel, which allows us to preserve local details present in the input image while inferring plausible ones in unseen regions.

Our end-to-end and unified digitization approach can directly predict high-resolution 3D shapes of a person with complex hairstyles and wearing arbitrary clothing. Despite the amount of unseen regions, particularly for a single-view input, our method can generate a complete model similar to ones obtained from multi-view stereo photogrammetry or other 3D scanning techniques. As shown in Figure \ref{fig:teaser}, our algorithm can handle a wide range of complex clothing, such as skirts, scarfs, and even high-heels while capturing high frequency details such as wrinkles that match the input image at the pixel level. 

By simply adopting the implicit function to regress RGB values at each queried point along the ray, PIFu can be naturally extended to infer per-vertex colors. Hence, our digitization framework also generates a complete texture of the surface, while predicting plausible appearance details in unseen regions. 
Through additional multi-view stereo constraints, PIFu can also be naturally extended to handle multiple input images, as is often desired for practical human capture settings. Since producing a complete textured mesh is already possible from a single input image, adding more views only improves our results further by providing additional information for unseen regions.

We demonstrate the effectiveness and accuracy of our approach on a wide range of challenging real-world and unconstrained images of clothed subjects. We also show for the first time, high-resolution examples of monocular and textured 3D reconstructions of dynamic clothed human bodies reconstructed from a video sequence. We provide comprehensive evaluations of our method using ground truth 3D scan datasets obtained using high-end photogrammetry. We compare our method with prior work and demonstrate the state-of-the-art performance on a public benchmark for digitizing clothed humans.

\section{Related Work}
\label{sec:related_work}

\noindent\textbf{Single-View 3D Human Digitization.}
Single-view digitization techniques require strong priors due to the ambiguous nature of the problem. Thus, parametric models of human bodies and shapes~\cite{anguelov2005scape, loper2015smpl} are widely used for digitizing humans from input images. Silhouettes and other types of manual annotations~\cite{guan2009estimating,zhou2010parametric} are often used to initialize the fitting of a statistical body model to images. Bogo et al.~\cite{bogo2016keep} proposed a fully automated pipeline for unconstrained input data. Recent methods involve deep neural networks to improve the robustness of pose and shape parameters estimations for highly challenging images~\cite{hmrKanazawa17,pavlakos2018humanshape}. Methods that involve part segmentation as input \cite{lassner2017unite,omran2018nbf} can produce more accurate fittings. Despite their capability to capture human body measurements and motions, parametric models only produce a naked human body. The 3D surfaces of clothing, hair, and other accessories are fully ignored. For skin-tight clothing, a displacement vector for each vertex is sometimes used to model some level of clothing as shown in~\cite{alldieck2018detailed,weng2018photo,alldieck2019learning}. Nevertheless, these techniques fail for more complex topology such as dresses, skirts, and long hair. To address this issue, template-free methods such as BodyNet \cite{varol18_bodynet} learn to directly generate a voxel representation of the person using a deep neural network. Due to the high memory requirements of voxel representations, fine-scale details are often missing in the output. More recently, \cite{natsume2018siclope} introduced a multi-view inference approach by synthesizing novel silhouette views from a single image. While multi-view silhouettes are more memory efficient, concave regions are difficult to infer as well as consistently generated views. Consequentially, fine-scale details cannot be produced reliably. In contrast, PIFu is memory efficient and is able to capture fine-scale details present in the image, as well as predict per-vertex colors.

\noindent\textbf{Multi-View 3D Human Digitization.}
Multi-view acquisition methods are designed to produce a complete model of a person and simplify the reconstruction problem, but are often limited to studio settings and calibrated sensors. Early attempts are based on visual hulls~\cite{matusik2000image, vlasic2008articulated,furukawa2006carved,esteban2004silhouette} which uses silhouettes from multiple views to carve out the visible areas of a capture volume. Reasonable reconstructions can be obtained when large numbers of cameras are used, but concavities are inherently challenging to handle. More accurate geometries can be obtained using multi-view stereo constraints~\cite{starck2007surface,zitnick2004high,waschbusch2005scalable,furukawa2010accurate} or using controlled illumination, such as multi-view photometric stereo techniques~\cite{vlasic2009dynamic,wu2012full}. Several methods use parametric body models to further guide the digitization process~\cite{sminchisescu2002human,gall2009motion,balan2007detailed,huang2017towards,alldieck2018video,alldieck2019learning}. The use of motion cues has also been introduced as additional priors~\cite{pons2017clothcap,yang2016estimation}. While it is clear that multi-view capture techniques outperform single-view ones, they are significantly less flexible and deployable.

A middle ground solution consists of using deep learning frameworks to generate plausible 3D surfaces from very sparse views. \cite{choy20163d} train a 3D convolutional LSTM to predict the 3D voxel representation of objects from arbitrary views. \cite{kar2017learning} combine information from arbitrary views using differentiable unprojection operations. \cite{ji2017surfacenet} also uses a similar approach, but requires at least two views. All of these techniques rely on the use of voxels, which is memory intensive and prevents the capture of high-frequency details. 
\cite{Huang18ECCV,gilbert:eccv:2018} introduced a deep learning approach based on a volumetric occupancy field that can capture dynamic clothed human performances using sparse viewpoints as input. At least three views are required for these methods to produce reasonable output.

\noindent\textbf{Texture Inference.}
When reconstructing a 3D model from a single image, the texture can be easily sampled from the input. However, the appearance in occluded regions needs to be inferred in order to obtain a complete texture.
Related to the problem of 3D texture inference are view-synthesis approaches that predict novel views from a single image \cite{zhou2016view,tvsn_cvpr2017} or multiple images \cite{sun2018multiview}. Within the context of texture mesh inference of clothed human bodies,~\cite{natsume2018siclope} introduced a view synthesis technique that can predict the back view from the front one. Both front and back views are then used to texture the final 3D mesh, however self-occluding regions and side views cannot be handled. Akin to the image inpainting problem \cite{pathakCVPR16context}, \cite{neverova2018dense} inpaints UV images that are sampled from the output of detected surface points, and \cite{drcTulsiani17,hspHane17} infers per voxel colors, but the output resolution is very limited.
\cite{cmrKanazawa18} directly predicts RGB values on a UV parameterization, but their technique can only handle shapes with known topology and are therefore not suitable for clothing inference.
Our proposed method can predict per vertex colors in an end-to-end fashion and can handle surfaces with arbitrary topology.

\begin{figure*}[t]
\begin{center}
  \includegraphics[width=0.8\linewidth]{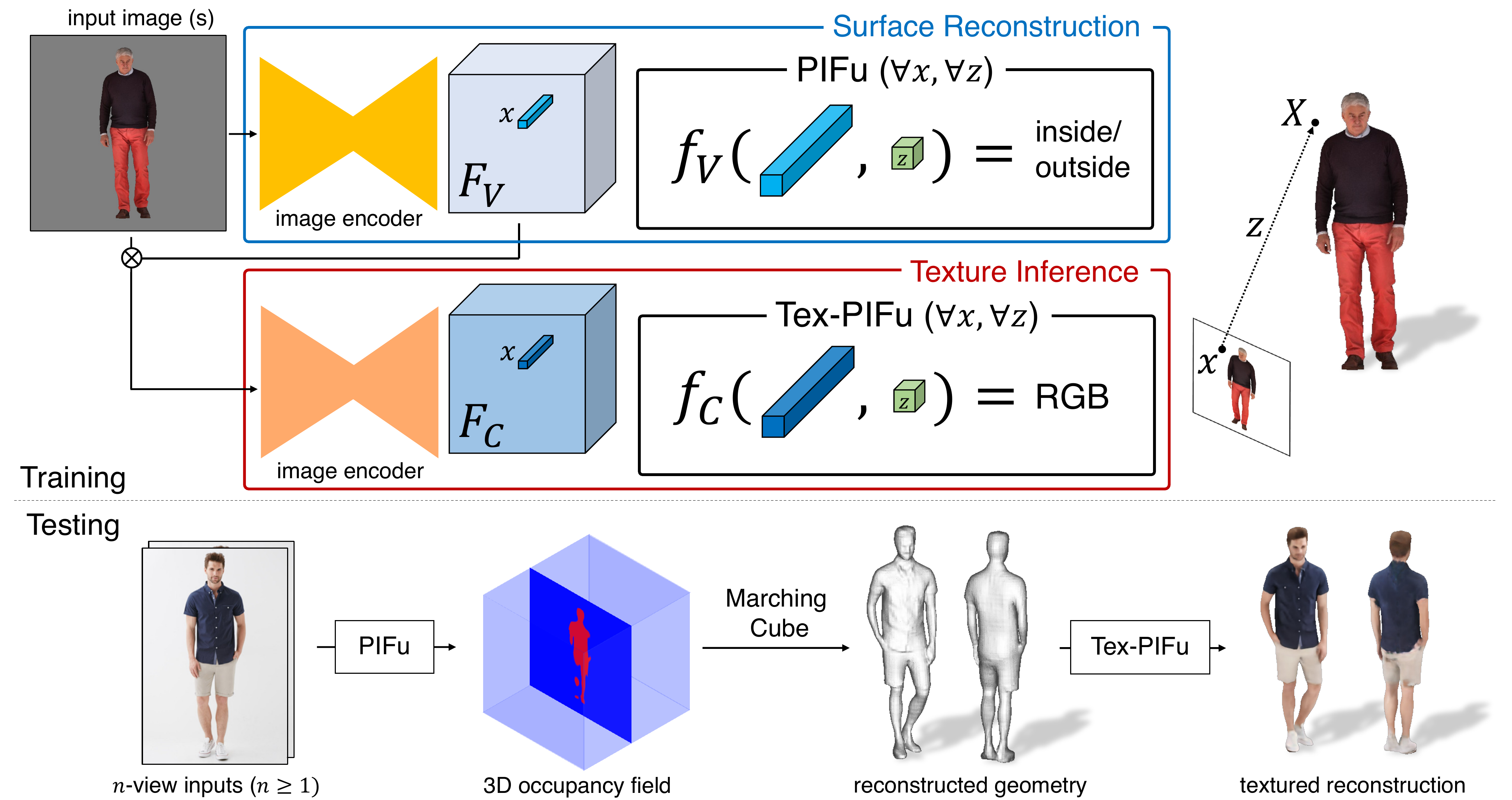}
\end{center}
\vspace{-10pt}
\caption{\textbf{Overview of our clothed human digitization pipeline:} Given an input image, a pixel-aligned implicit function (PIFu) predicts the continuous inside/outside probability field of a clothed human. Similarly, PIFu for texture inference (Tex-PIFu) infers an RGB value at given 3D positions of the surface geometry with arbitrary topology.
}
\vspace{-15pt}
\label{fig:overview}
\end{figure*}
\section{PIFu: Pixel-Aligned Implicit Function}
\label{sec:method}
Given a single or multi-view images, our goal is to reconstruct the underlining 3D geometry and texture of a clothed human while preserving the detail present in the image.
To this end, we introduce Pixel-Aligned Implicit Functions (PIFu) which is a memory efficient and spatially-aligned 3D representation for 3D surfaces. An implicit function defines a surface as a level set of a function $f$, \eg $f(X) = 0$ \cite{sclaroff1991generalized}. This results in a memory efficient representation of a surface where the space in which the surface is embedded does not need to be explicitly stored. The proposed pixel-aligned implicit function consists of a fully convolutional image encoder $g$ and a continuous implicit function $f$ represented by multi-layer perceptrons (MLPs), where the surface is defined as a level set of
\begin{equation}
\label{eq:PIFu}
f(F(x), z(X)) = s : s \in \mathbb{R},
\end{equation}
where for a 3D point $X$, $x=\pi(X)$ is its 2D projection, $z(X)$ is the depth value in the camera coordinate space, $F(x) = g(I(x))$ is the image feature at $x$. We assume a weak-perspective camera, but extending to perspective cameras is straightforward. Note that we obtain the pixel-aligned feature $F(x)$ using bilinear sampling, because the 2D projection of $X$ is defined in a continuous space rather than a discrete one (i.e., pixel). 

The key observation is that we learn an implicit function over the 3D space with pixel-aligned image features rather than global features, which allows the learned functions to preserve the local detail present in the image. The continuous nature of PIFu allows us to generate detailed geometry with arbitrary topology in a memory efficient manner. Moreover, PIFu can be cast as a general framework that can be extended to various co-domains such as RGB colors. 

\vspace{-5pt}
\paragraph{Digitization Pipeline.}
Figure \ref{fig:overview} illustrates the overview of our framework. Given an input image, PIFu for surface reconstruction 
predicts the continuous inside/outside probability field of a clothed human, in which iso-surface can be easily extracted (Sec. \ref{sec:v-PIFu}). Similarly, PIFu for texture inference (Tex-PIFu) outputs an RGB value at 3D positions of the surface geometry, enabling texture inference in self-occluded surface regions and shapes of arbitrary topology (Sec. \ref{sec:c-PIFu}). Furthermore, we show that the proposed approach can handle single-view and multi-view input naturally, which allows us to produce even higher fidelity results when more views are available (Sec. \ref{sec:mv-PIFu}).

\subsection{Single-view Surface Reconstruction}
\label{sec:v-PIFu}
For surface reconstruction, we represent the ground truth surface as a $0.5$ level-set of a continuous 3D occupancy field:
\begin{equation}
\label{eq:sdf_def}
    f^*_v(X) = 
    \begin{cases}
        1,& \text{if } X \text{ is inside mesh surface}\\
        0, & \text{otherwise}
    \end{cases}.
\end{equation}
We train a pixel-aligned implicit function (PIFu) $f_v$ by minimizing the average of mean squared error:
\begin{equation}
\label{eq:v_loss}
\mathcal{L}_V = \frac{1}{n}\sum_{i=1}^n{\left|f_v(F_V(x_i),z(X_i)) - f^*_v(X_i)\right|^2},
\end{equation}
where $X_i \in \mathbb{R}^3$, $F_V(x) = g(I(x))$ is the image feature from the image encoder $g$ at $x=\pi(X)$ and $n$ is the number of sampled points. Given a pair of an input image and the corresponding 3D mesh that is spatially aligned with the input image, the parameters of the image encoder $g$ and PIFu $f_v$ are jointly updated by minimizing Eq. \ref{eq:v_loss}. As Bansal et al. \cite{PixelNet} demonstrate for semantic segmentation, training an image encoder with a subset of pixels does not hurt convergence compared with training with all the pixels.
During inference, we densely sample the probability field over the 3D space and extract the iso-surface of the probability field at threshold $0.5$ using the Marching Cube algorithm \cite{lorensen1987marching}. This implicit surface representation is suitable for detailed objects with arbitrary topology. Aside from PIFu's expressiveness and memory-efficiency, we develop a spatial sampling strategy that is critical for achieving high-fidelity inference.

\paragraph{Spatial Sampling.}
The resolution of the training data plays a central role in achieving the expressiveness and accuracy of our implicit function. Unlike voxel-based methods, our approach does not require discretization of ground truth 3D meshes. Instead, we can directly sample 3D points on the fly from the ground truth mesh in the original resolution using an efficient ray tracing algorithm \cite{wald2014embree}. Note that this operation requires water-tight meshes. In the case of non-watertight meshes, one can use off-the-shelf solutions to make the meshes watertight \cite{Barill:FW:2018}. Additionally, we observe that the sampling strategy can largely influence the final reconstruction quality. If one uniformly samples points in the 3D space, the majority of points are far from the iso-surface, which would unnecessarily weight the network toward outside predictions. On the other hand, sampling only around the iso-surface can cause overfitting. Consequently, we propose to combine uniform sampling and adaptive sampling based on the surface geometry. We first randomly sample points on the surface geometry and add offsets with normal distribution $\mathcal{N}(0,\sigma)$ ($\sigma=5.0$ cm in our experiments) for x, y, and z axis to perturb their positions around the surface. We combine those samples with uniformly sampled points within bounding boxes using a ratio of $16:1$. We provide an ablation study on our sampling strategy in the supplemental materials.

\subsection{Texture Inference}
\label{sec:c-PIFu}
While texture inference is often performed on either a 2D parameterization of the surface \cite{cmrKanazawa18,guler2018densepose} or in view-space \cite{natsume2018siclope}, PIFu enables us to directly predict the RGB colors on the surface geometry by defining $s$ in Eq. \ref{eq:PIFu} as an RGB vector field instead of a scalar field. This supports texturing of shapes with arbitrary topology and self-occlusion. 
However, extending PIFu to color prediction is a non-trivial task as RGB colors are defined only on the surface while the 3D occupancy field is defined over the entire 3D space. Here, we highlight the modification of PIFu in terms of training procedure and network architecture. 

Given sampled 3D points on the surface $X \in \Omega$, the objective function for texture inference is the average of L1 error of the sampled colors as follows:
\begin{equation}
\label{eq:c_loss}
\mathcal{L}_C = \frac{1}{n}\sum_{i=1}^n{\left|f_c(F_C(x_i),z(X_i)) - C(X_i)\right|},
\end{equation}
where $C(X_i)$ is the ground truth RGB value on the surface point $X_i \in \Omega$ and $n$ is the number of sampled points. We found that naively training $f_c$ with the loss function above severely suffers from overfitting. The problem is that $f_c$ is expected to learn not only RGB color on the surface but also the underlining 3D surfaces of the object so that $f_c$ can infer texture of unseen surface with different pose and shape during inference, which poses a significant challenge. We address this problem with the following modifications. First, we condition the image encoder for texture inference with the image features learned for the surface reconstruction $F_V$. This way, the image encoder can focus on color inference of a given geometry even if unseen objects have different shape, pose, or topology. Additionally, we introduce an offset $\epsilon \sim \mathcal{N}(0, d)$ to the surface points along the surface normal $N$ so that the color can be defined not only on the exact surface but also on the 3D space around it. With the modifications above, the training objective function can be rewritten as:
\begin{equation}
\label{eq:c_loss_mod}
\mathcal{L}_C = \frac{1}{n}\sum_{i=1}^n{\left|f_c(F_C(x'_i, F_V),X'_{i,z}) - C(X_i)\right|},
\end{equation}
where $X'_i = X_i + \epsilon\cdot N_i$. We use $d=1.0$ cm for all the experiments. Please refer to the supplemental material for the network architecture for texture inference.

\begin{figure}[t]
\begin{center}
  \includegraphics[width=\linewidth]{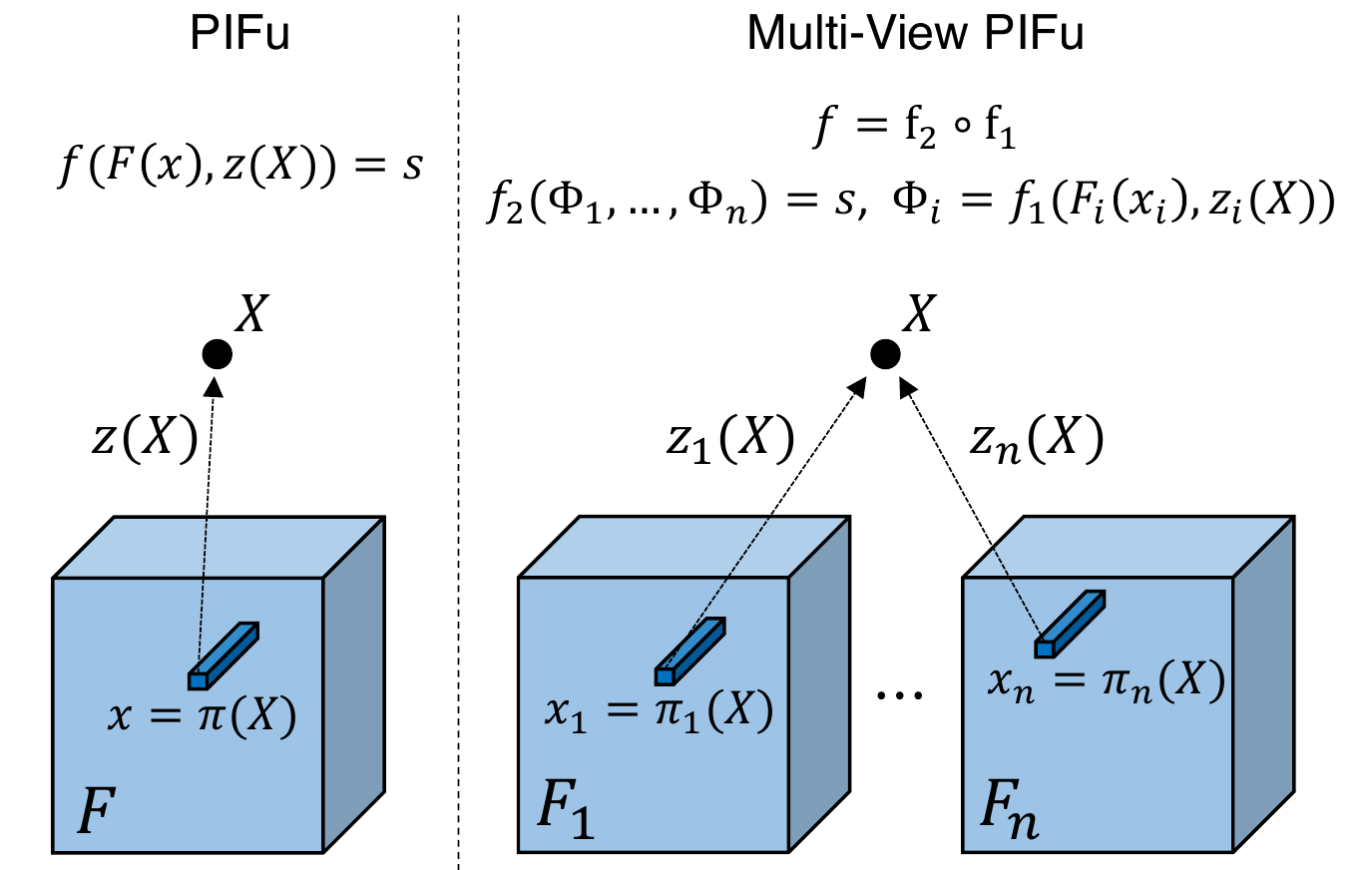}
\end{center}
\vspace{-15pt}
\caption{\textbf{Multi-view PIFu:}
PIFu can be extended to support multi-view inputs by decomposing implicit function $f$ into a feature embedding function $f_1$ and a multi-view reasoning function $f_2$. $f_1$ computes a feature embedding from each view in the 3D world coordinate system, which allows aggregation from arbitrary views. $f_2$ takes aggregated feature vector to make a more informed 3D surface and texture prediction. 
}
\vspace{-9pt}

\label{fig:mv_pifu}
\end{figure}
\subsection{Multi-View Stereo} 
\label{sec:mv-PIFu}
Additional views provide more coverage about the person and should improve the digitization accuracy. Our formulation of PIFu provides the option to incorporate information from more views for both surface reconstruction and texture inference. We achieve this by using PIFu to learn a feature embedding for every 3D point in space. Specifically the output domain of Eq.~\ref{eq:PIFu} is now a $n$-dimensional vector space $s \in \mathbb{R}^n$ that represents the latent feature embedding associated with the specified 3D coordinate and the image feature from each view. Since this embedding is defined in the 3D world coordinate space, we can aggregate the embedding from all available views that share the same 3D point. The aggregated feature vector can be used to make a more confident prediction of the surface and the texture.

Specifically we decompose the pixel-aligned function $f$ into a feature embedding network $f_1$ and a multi-view reasoning network $f_2$ as $f := f_2 \circ f_1$. See Figure \ref{fig:mv_pifu} for illustrations. 
The first function $f_1$ encodes the image feature $F_i(x_i): x_i = \pi_i(X)$ and depth value $z_i(X)$ from each view point $i$ into latent feature embedding $\Phi_i$. This allows us to aggregate the corresponding pixel features from all the views. Now that the corresponding 3D point $X$ is shared by different views, each image can project $X$ on its own image coordinate system by $\pi_i(X)$ and $z_i(X)$. Then, we aggregate the latent features $\Phi_i$ by average pooling operation and obtain the fused embedding $\bar{\Phi}=\text{mean}(\{\Phi_i\})$. The second function $f_2$ maps from the aggregated embedding $\bar{\Phi}$ to our target implicit field $s$ (i.e., inside/outside probability for surface reconstruction and RGB value for texture inference).  The additive nature of the latent embedding allows us to incorporate arbitrary number of inputs. Note that a single-view input can be also handled without modification in the same framework as the average operation simply returns the original latent embedding. 
For training, we use the same training procedure as the aforementioned single-view cases including loss functions and the point sampling scheme. While we train with three random views, our experiments show that the model can incorporate information from more than three views (See Sec. \ref{sec:evaluation}). 
\begin{figure*}[t]
\begin{center}
  \includegraphics[width=0.75\textwidth]{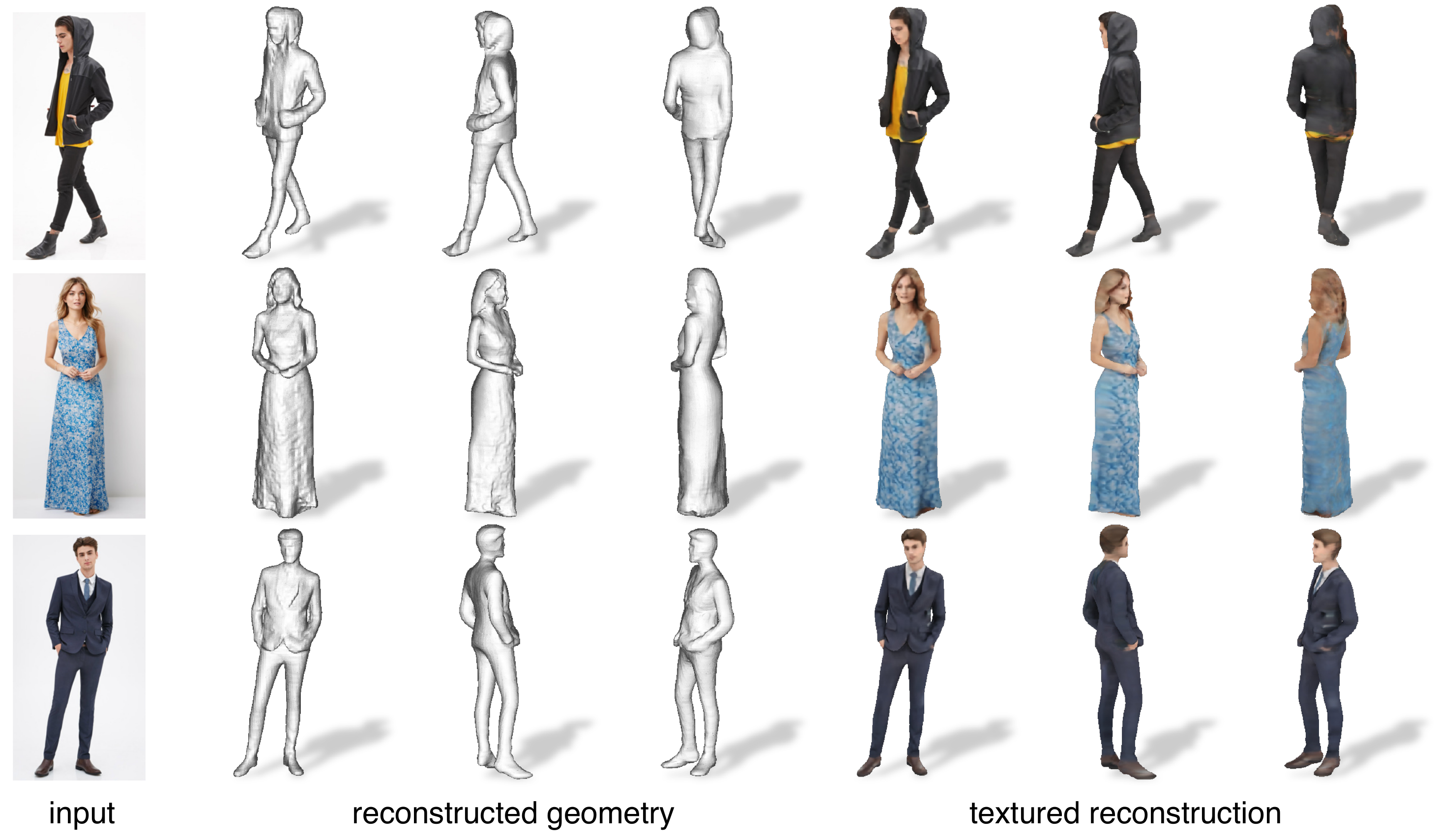}
\end{center}
\vspace{-.5cm}
\caption{\textbf{Qualitative single-view results on real images from DeepFashion dataset~\cite{liu2016deepfashion}.} The proposed Pixel-Aligned Implicit Functions, PIFu, achieves a topology-free,  memory efficient, spatially-aligned 3D reconstruction of geometry and texture of clothed human.}
\label{fig:big}
\end{figure*}

\section{Experiments}
\label{sec:result}

\label{sec:evaluation}
We evaluate our proposed approach on a variety of datasets,
including RenderPeople \cite{renderpeople} and BUFF \cite{zhang-CVPR17}, which has ground truth measurements, as well as DeepFashion \cite{liu2016deepfashion} which contains a diverse variety of complex clothing.

\paragraph{Implementation Detail.}
Since the framework of PIFu is not limited to a specific network architecture, one can technically use any fully convolutional neural network as the image encoder. For surface reconstruction, we found that stacked hourglass \cite{newell2016stacked} architectures are effective with better generalization on real images. The image encoder for texture inference adopts the architecture of CycleGAN \cite{zhu2017unpaired} consisting of residual blocks \cite{johnson2016perceptual}. The implicit function is based on a multi-layer perceptron, whose layers have skip connections from the image feature $F(x)$ and depth $z$ in spirit of \cite{chen2018implicit_decoder} to effectively propagate the depth information. Tex-PIFu takes $F_C(x)$ together with the image feature for surface reconstruction $F_V(x)$ as input. For multi-view PIFu, we simply take an intermediate layer output as feature embedding and apply average pooling to aggregate the embedding from different views. Please refer to the supplemental materials for more detail on network architecture and training procedure.

\subsection{Quantitative Results}
We quantitatively evaluate our reconstruction accuracy with three metrics. In the model space, we measure the average point-to-surface Euclidean distance (P2S) in cm from the vertices on the reconstructed surface to the ground truth. We also measure the Chamfer distance between the reconstructed and the ground truth surfaces. In addition, we introduce the normal reprojection error to measure the fineness of reconstructed local details, as well as the projection consistency from the input image.
For both reconstructed and ground truth surfaces, we render their normal maps in the image space from the input viewpoint respectively.
We then calculate the L$2$ error between these two normal maps.

\begin{figure*}[ht]
\begin{center}
  \includegraphics[width=0.80\textwidth,trim={0 0 0 8.5cm},clip]{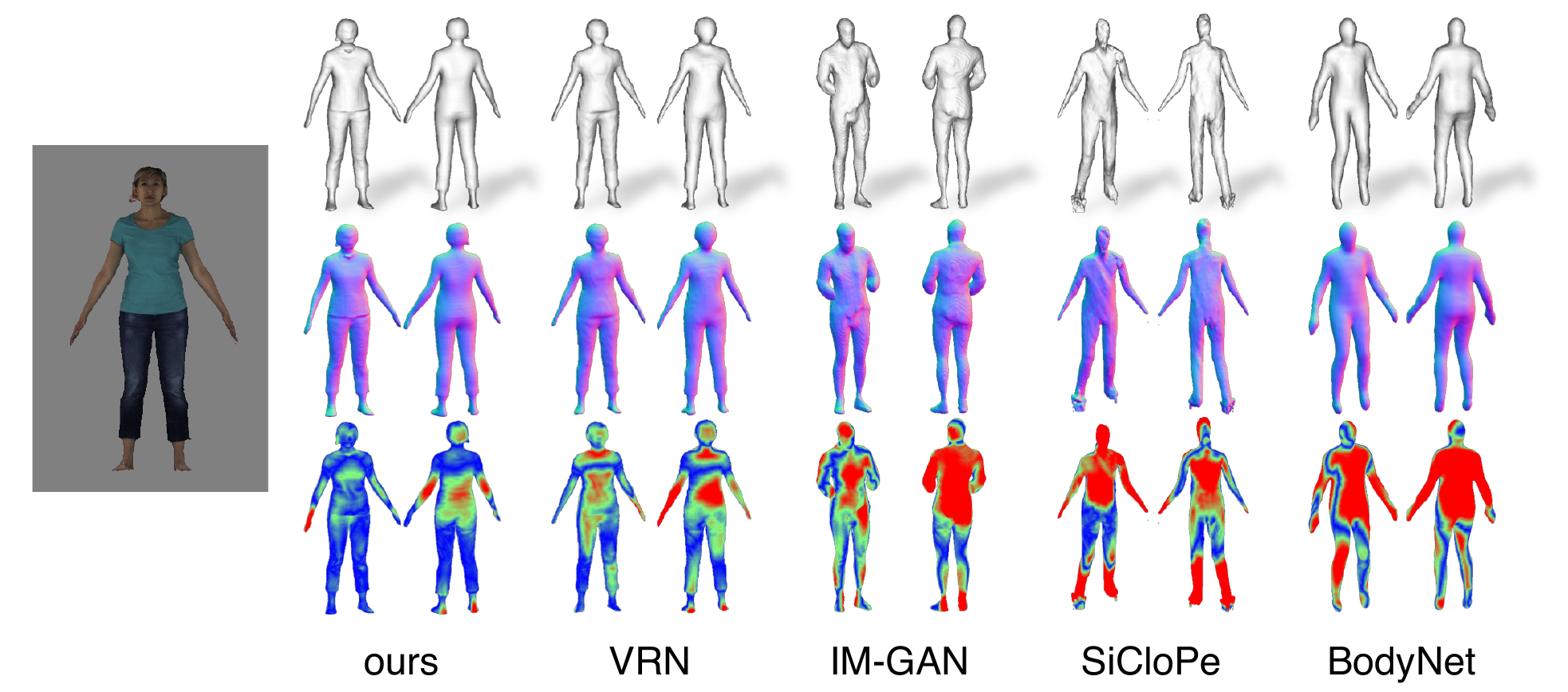}
\end{center}
\vspace{-10pt}
\caption{\textbf{Comparison with other human digitization methods from a single image.} For each input image on the left, we show the predicted surface (top row), surface normal (middle row), and the point-to-surface errors (bottom row). }
\label{fig:single}
\end{figure*}

\noindent\textbf{Single-View Reconstruction.}
In Table \ref{tab:main} and Figure \ref{fig:single},
we evaluate the reconstruction errors for each method on both 
Buff and RenderPeople test set. Note that while Voxel Regression Network (VRN) \cite{jackson2018human}, IM-GAN \cite{chen2018implicit_decoder}, and ours are retrained with the same High-Fidelity Clothed Human dataset we use for our approach, the reconstruction of \cite{natsume2018siclope, varol18_bodynet} are obtained from their trained models as off-the-shelf solutions. Since single-view inputs leaves the scale factor ambiguous, the evaluation is performed with the known scale factor for all the approaches. In contrast to the state-of-the-art single-view reconstruction method using implicit function (IM-GAN) \cite{chen2018encoder} that reconstruct surface from one global feature per image, our method outputs pixel-aligned high-resolution surface reconstruction that captures hair styles and wrinkles of the clothing. We also demonstrate the expressiveness of our PIFu representation compared with voxels. Although VRN and ours share the same network architecture for the image encoder, the higher expressiveness of implicit representation allows us to achieve higher fidelity. 

\begin{figure}[t]
\begin{center}
  \includegraphics[width=\linewidth]{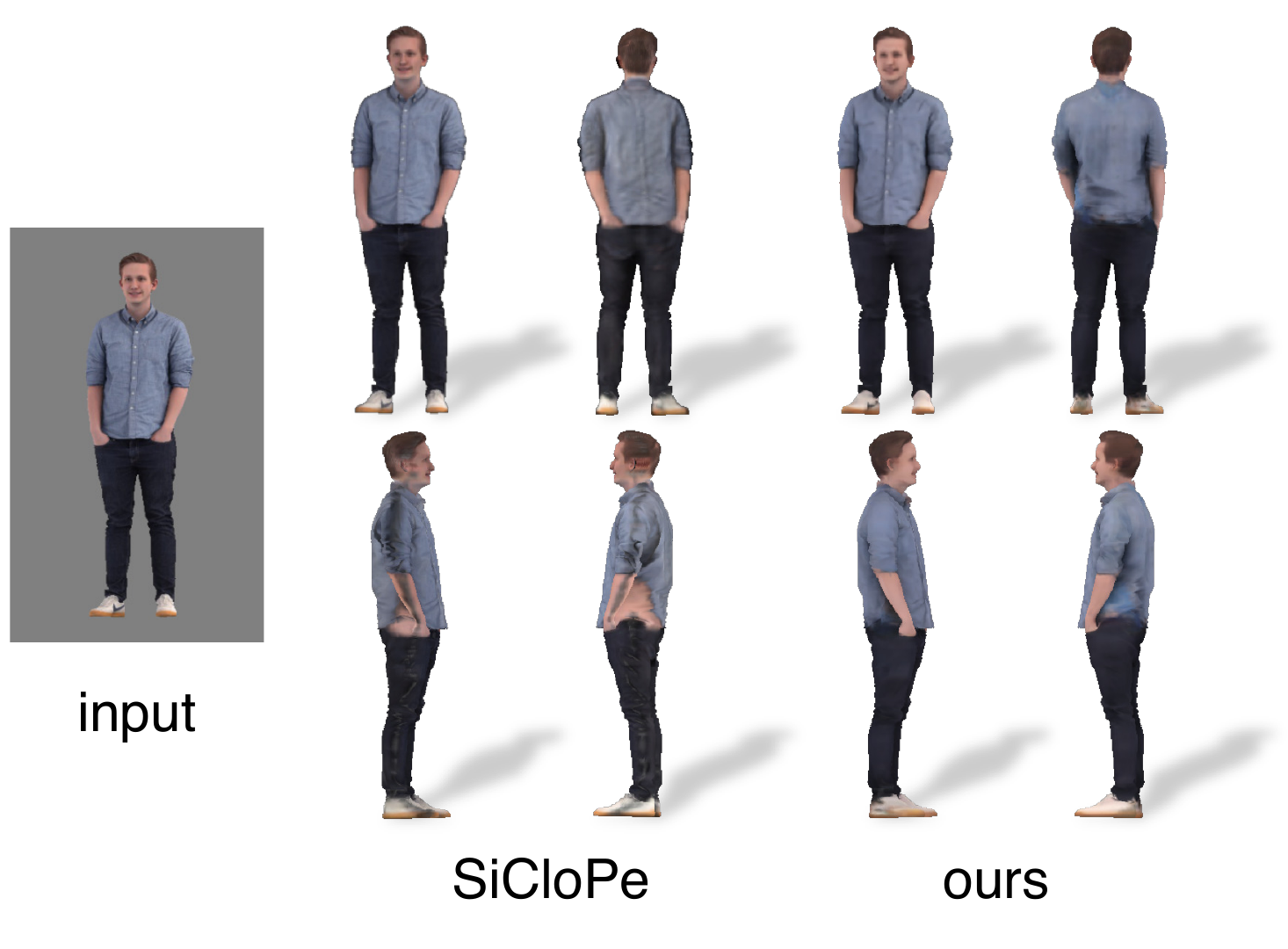}
\end{center}
\vspace{-20pt}
\caption{\small{
\textbf{Comparison with SiCloPe \cite{natsume2018siclope} on texture inference.} While texture inference via a view synthesis approach suffers from projection artifacts, proposed approach does not as it directly inpaints textures on the surface geometry.
}}
\vspace{-10pt}
\label{fig:color_comp}
\end{figure}

In Figure \ref{fig:color_comp}, we also compare our single-view texture inferences with a state-of-the-art texture inference method on clothed human, SiCloPe \cite{natsume2018siclope}, which infers a 2D image from the back view and stitches it together with the input front-view image to obtain textured meshes. While SiCloPe suffers from projection distortion and artifacts around the silhouette boundary, our approach predicts textures on the surface mesh directly, removing projection artifacts.

\begin{figure}[t]
\begin{center}
  \includegraphics[width=\linewidth]{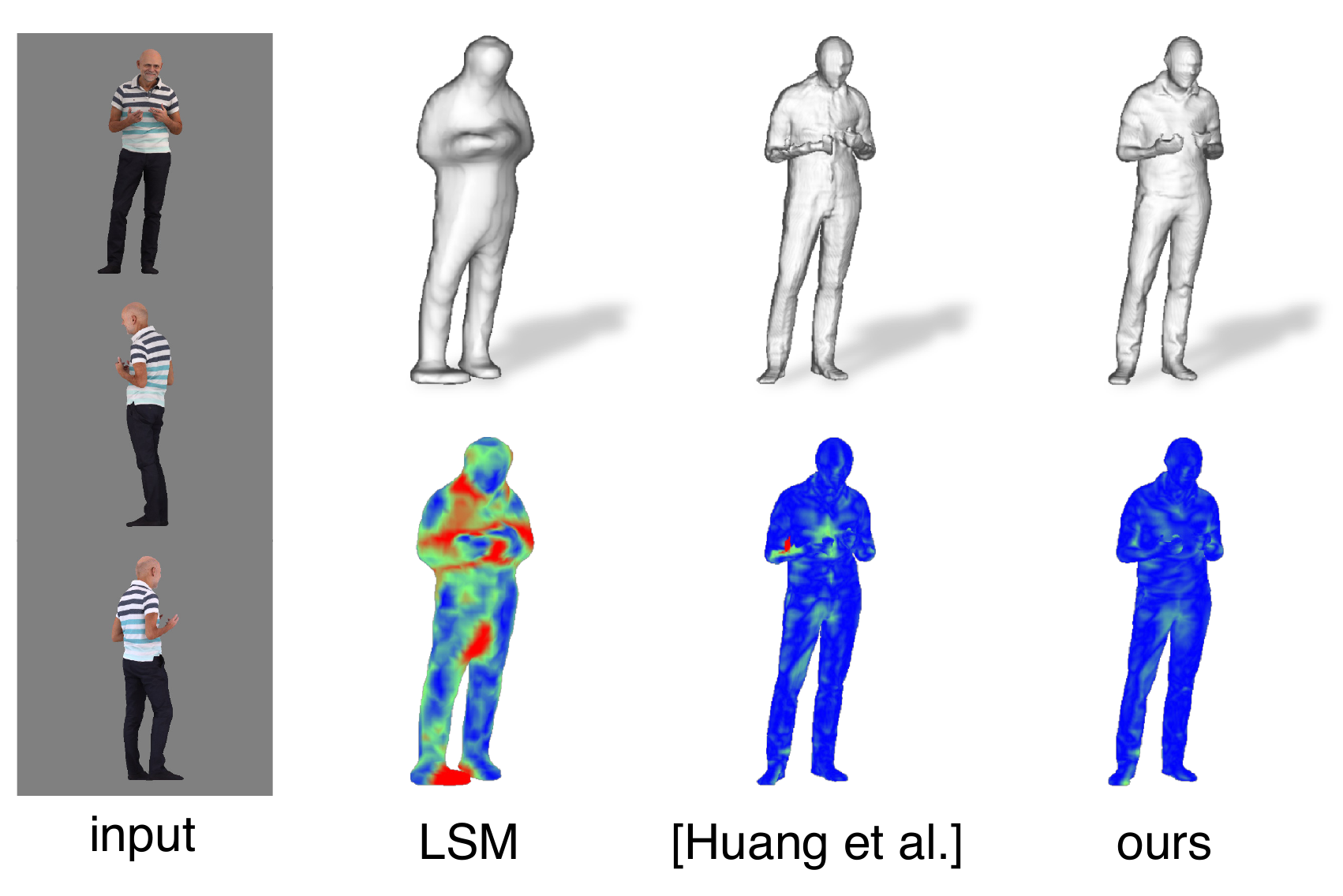}
\end{center}
\vspace{-15pt}
\caption{\small{
\textbf{Comparison with learning-based multi-view methods.} Ours outperforms other learning-based multi-view methods qualitatively and quantitatively. Note that all methods are trained with three view inputs from the same training data.
}}
\vspace{-5pt}
\label{fig:mv_comp}
\end{figure}

\begin{figure}[t]
\begin{center}
  \includegraphics[width=\linewidth]{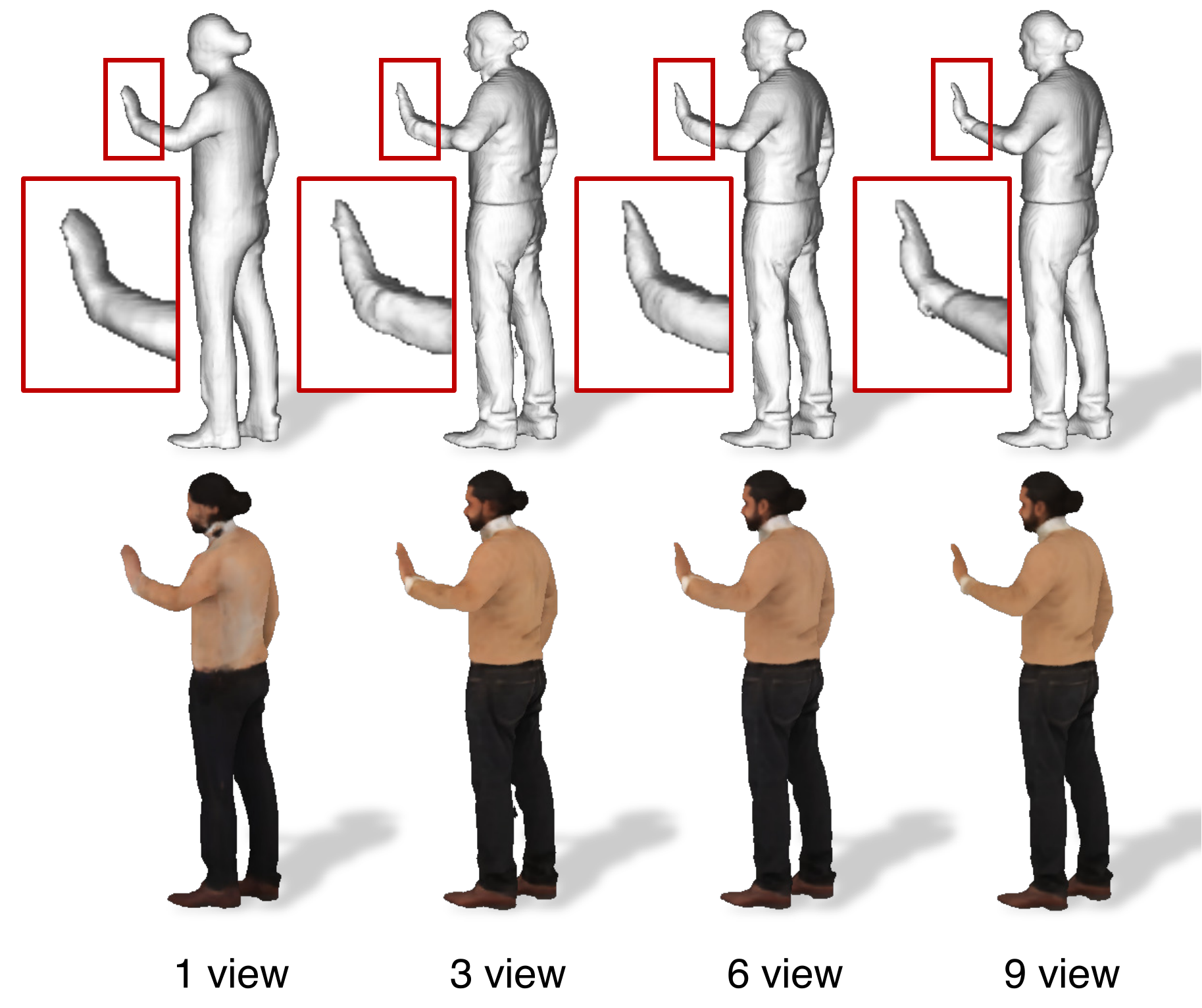}
\end{center}
\vspace{-15pt}
\caption{
Our surface and texture predictions increasingly improve as more views are added.}
\vspace{-15pt}
\label{fig:mv_vis}
\end{figure}

\noindent\textbf{Multi-View Reconstruction.}
In Table \ref{tab:multiview} and Figure \ref{fig:mv_comp}, we compare our multi-view reconstruction with other deep learning-based multi-view methods including LSM \cite{kar2017learning}, and a deep visual hull method proposed by Huang et al. \cite{huang2018learning}. All approaches are trained on the same High-Fidelity Clothed Human Dataset using three-view input images. Note that Huang et al. can be seen as a degeneration of our method where the multi-view feature fusion process solely relies on image features, without explicit conditioning on the 3D coordinate information. To evaluate the importance of conditioning on the depth, we denote our network architecture removing $z$ from input of PIFu as Huang et al. in our experiments. We demonstrate that PIFu achieves the state-of-the-art reconstruction qualitatively and quantitatively in our metrics. We also show that our multi-view PIFu allows us to increasingly refine the geometry and texture by incorporating arbitrary number of views in Figure \ref{fig:mv_vis}. 

\subsection{Qualitative Results}
In Figure \ref{fig:big}, we present our digitization results using real world input images from the DeepFashion dataset \cite{liu2016deepfashion}. We demonstrate our PIFu can handle wide varieties of clothing, including skirts, jackets, and dresses. Our method can produce high-resolution local details, while inferring plausible 3D surfaces in unseen regions. Complete textures are also inferred successfully from a single input image, which allows us to view our 3D models from $360$ degrees. We refer to the supplemental video\footnote{\url{https://youtu.be/S1FpjwKqtPs}} for additional static and dynamic results. In particular, we show how dynamic clothed human performances and complex deformations can be digitized in 3D from a single 2D input video.

\begin{table}[t]
\resizebox{\columnwidth}{!}{
\begin{tabular}{llll|lll}
\hline
           & \multicolumn{3}{c|}{RenderPeople} & \multicolumn{3}{c}{Buff} \\
Methods    & Normal     & P2S     & Chamfer    & Normal  & P2S         & Chamfer \\ \hline
BodyNet    & 0.262      & 5.72    & 5.64       & 0.308   & 4.94        & 4.52    \\
SiCloPe    & 0.216      & 3.81    & 4.02        & 0.222  & 4.06         & 3.99        \\ 
IM-GAN     & 0.258      & 2.87    & 3.14       & 0.337   & 5.11        & 5.32    \\
VRN        & 0.116      & \textbf{1.42}    & 1.56       & 0.130   & 2.33        & 2.48   \\
Ours       & \textbf{0.084}      & 1.52    & \textbf{1.50}       & \textbf{0.0928}  & \textbf{1.15}        & \textbf{1.14}
\end{tabular}
}
\caption{Quantitative evaluation on RenderPeople and BUFF dataset for single-view reconstruction.}
\label{tab:main}
\end{table}

\begin{table}[t]
\resizebox{\columnwidth}{!}{
\begin{tabular}{llll|lll}
\hline
           & \multicolumn{3}{c|}{RenderPeople} & \multicolumn{3}{c}{Buff} \\
Methods     & Normal     & P2S       & Chamfer      & Normal    & P2S           & Chamfer \\ \hline
LSM         & 0.251     & 4.40      & 3.93          & 0.272     & 3.58          & 3.30 \\ 
Deep V-Hull & \textbf{0.093}      & 0.639     & 0.632        & 0.119      & 0.698        & 0.709  \\
Ours        & 0.094      & \textbf{0.554}     & \textbf{0.567}        & \textbf{0.107}      & \textbf{0.665}        & \textbf{0.641}
\end{tabular}
}
\caption{Quantitative comparison between multi-view reconstruction algorithms using $3$ views.}
\label{tab:multiview}
\end{table}

\section{Discussion}
\label{sec:conclusion}

We introduced a novel pixel-aligned implicit function, which  spatially aligns the pixel-level information of the input image with the shape of the 3D object, for deep learning based 3D shape and texture inference of clothed humans from a single input image. Our experiments indicate that highly plausible geometry can be inferred including largely unseen regions such as the back of a person, while preserving high-frequency details present in the image. Unlike voxel-based representations, our method can produce high-resolution output since we are not limited by the high memory requirements of volumetric representations. Furthermore, we also demonstrate how this method can be naturally extended to infer the entire texture on a person given partial observations. Unlike existing methods, which synthesize the back regions based on frontal views in an image space, our approach can predict colors in unseen, concave and side regions directly on the surface. In particular, our method is the first approach that can inpaint textures for shapes of arbitrary topology. Since we are capable for generating textured 3D surfaces of a clothed person from a single RGB camera, we are moving a step closer toward monocular reconstructions of dynamic scenes from video without the need of a template model. Our ability to handle arbitrary additional views also makes our approach particularly suitable for practical and efficient 3D modeling settings using sparse views, where traditional multi-view stereo or structure-from-motion would fail. 

\paragraph{Future Work.} While our texture predictions are reasonable and not limited by the topology or parameterization of the inferred 3D surface, we believe that higher resolution appearances can be inferred, possibly using generative adversarial networks or increasing the input image resolution. In this work, the reconstruction takes place in pixel coordinate space, aligning the scale of subjects as pre-process. As in other single-view methods, inferring scale factor remains an open-question, which future work can address. Lastly, in all our examples, none of the segmented subjects are occluded by any other objects or scene elements. In real-world settings, occlusions often occur and perhaps only a part of the body is framed in the camera. Being able to digitize and predict complete objects in partially visible settings could be highly valuable for analyzing humans in unconstrained settings. 
\vspace{1em}

\small{
\noindent\textbf{Acknowledgements}
Hao Li is affiliated with the University of Southern California, the USC Institute for Creative Technologies, and Pinscreen. This research was conducted at USC and was funded by in part by the ONR YIP grant N00014-17-S-FO14, the CONIX Research Center, one of six centers in JUMP, a Semiconductor Research Corporation program sponsored by DARPA, the Andrew and Erna Viterbi Early Career Chair, the U.S. Army Research Laboratory under contract number W911NF-14-D-0005, Adobe, and Sony. This project was not funded by Pinscreen, nor has it been conducted at Pinscreen or by anyone else affiliated with Pinscreen. Shigeo Morishima is supported by the JST ACCEL Grant Number JPMJAC1602, JSPS KAKENHI Grant Number JP17H06101, JP19H01129. Angjoo Kanazawa is supported by BAIR sponsors. The content of the information does not necessarily reflect the position or the policy of the Government, and no official endorsement should be inferred.
}
{\small
	\bibliographystyle{ieee_fullname}
	\bibliography{paper}
}
\section*{Appendix I. Implementation Details}

\paragraph{Experimental Setup.}
Since there is no large scale datasets for high-resolution clothed humans, we collected photogrammetry data of $491$ high-quality textured human meshes with a wide range of clothing, shapes, and poses, each consisting of about $100,000$ triangles from RenderPeople\footnote{\url{https://renderpeople.com/3d-people/}}. We refer to this database as High-Fidelity Clothed Human Data set. We randomly split the dataset into a training set of $442$ subjects and a test set of $49$ subjects. 
To efficiently render the digital humans, Lambertian diffuse shading with surface normal and spherical harmonics are typically used due to its simplicity and efficiency \cite{varol17_surreal, natsume2018siclope}. However, we found that to achieve high-fidelity reconstructions on real images, the synthetic renderings need to correctly simulate light transport effects resulting from both global and local geometric properties such as ambient occlusion. To this end, we use a precomputed radiance transfer technique (PRT) that precomputes visibility on the surface using spherical harmonics and efficiently represents global light transport effects by multiplying spherical harmonics coefficients of illumination and visibility  \cite{sloan2002precomputed}. PRT only needs to be computed once per object and can be reused with arbitrary illuminations and camera angles. Together with PRT, we use $163$ second-order spherical harmonics of indoor scene from HDRI Haven\footnote{\url{https://hdrihaven.com/}} using random rotations around y axis. We render the images by aligning subjects to the image center using a weak-perspective camera model and image resolution of $512\times512$. We also rotate the subjects for $360$ degrees in yaw axis, resulting in $360\times442=159,120$ images for training. For the evaluation, we render $49$ subjects from RenderPeople and $5$ subjects from the BUFF data set \cite{zhang-CVPR17} using $4$ views spanning every $90$ degrees in yaw axis. Note that we render the images without the background. We also test our approach on real images of humans from the DeepFashion data set \cite{liu2016deepfashion}. In the case of real data, we use a off-the-shelf semantic segmentation network together with Grab-Cut refinement \cite{rother2004grabcut}.

\paragraph{Network Architecture.}
Since the framework of PIFu is not limited to a specific network architecture, one can technically use any fully convolutional neural network as the image encoder. For surface reconstruction, we adapt the stacked hourglass network \cite{newell2016stacked} with modifications proposed by \cite{jackson2018human}. We also replace batch normalization with group normalization \cite{wu2018group}, which improves the training stability when the batch sizes are small. Similar to \cite{jackson2018human}, the intermediate features of each stack are fed into PIFu, and the losses from all the stacks are aggregated for parameter update. We have conducted ablation study on the network architecture design and compare against other alternatives (VGG16, ResNet34) in Appendix II. The image encoder for texture inference adopts the architecture of CycleGAN \cite{zhu2017unpaired} consisting of $6$ residual blocks \cite{johnson2016perceptual}. Instead of using transpose convolutions to upsample the latent features, we directly feed the output of the residual blocks to the following Tex-PIFu.

PIFu for surface reconstruction is based on a multi-layer perceptron, where the number of neurons is $(257,1024,512,256,128,1)$ with non-linear activations using leaky ReLU except the last layer that uses sigmoid activation. To effectively propagate the depth information, each layer of MLP has skip connections from the image feature $F(x) \in \mathbb{R}^{256}$ and depth $z$ in spirit of \cite{chen2018implicit_decoder}. For multi-view PIFu, we simply take the $4$-th layer output as feature embedding and apply average pooling to aggregate the embedding from different views. Tex-PIFu takes $F_C(x) \in \mathbb{R}^{256}$ together with the image feature for surface reconstruction $F_V(x) \in \mathbb{R}^{256}$ by setting the number of the first neurons in the MLP to $513$ instead of $257$. We also replace the last layer of PIFu with $3$ neurons, followed by tanh activation to represent RGB values. 

\paragraph{Training procedure.}
Since the texture inference module requires pretrained image features from the surface reconstruction module, we first train PIFu for the surface reconstruction and then for texture inference, using the learned image features $F_V$ as condition. We use RMSProp for the surface reconstruction following \cite{newell2016stacked} and Adam for the texture inference with learning rate of $1\times10^{-3}$ as in \cite{zhu2017unpaired}, the batch size of $3$ and $5$, the number of epochs of $12$ and $6$, and the number of sampled points of $5000$ and $10000$ per object in every training batch respectively. The learning rate of RMSProp is decayed by the factor of $0.1$ at $10$-th epoch following \cite{newell2016stacked}. The multi-view PIFu is fine-tuned from the models trained for single-view surface reconstruction and texture inference with a learning rate of $1\times10^{-4}$ and $2$ epochs. The training of PIFu for single-view surface reconstruction and texture inference takes $4$ and $2$ days, respectively, and fine-tuning for multi-view PIFu can be achieved within $1$ day on a single 1080ti GPU.
\section*{Appendix II. Additional Evaluations}

\begin{figure}[t]
\begin{center}
  \includegraphics[width=\linewidth]{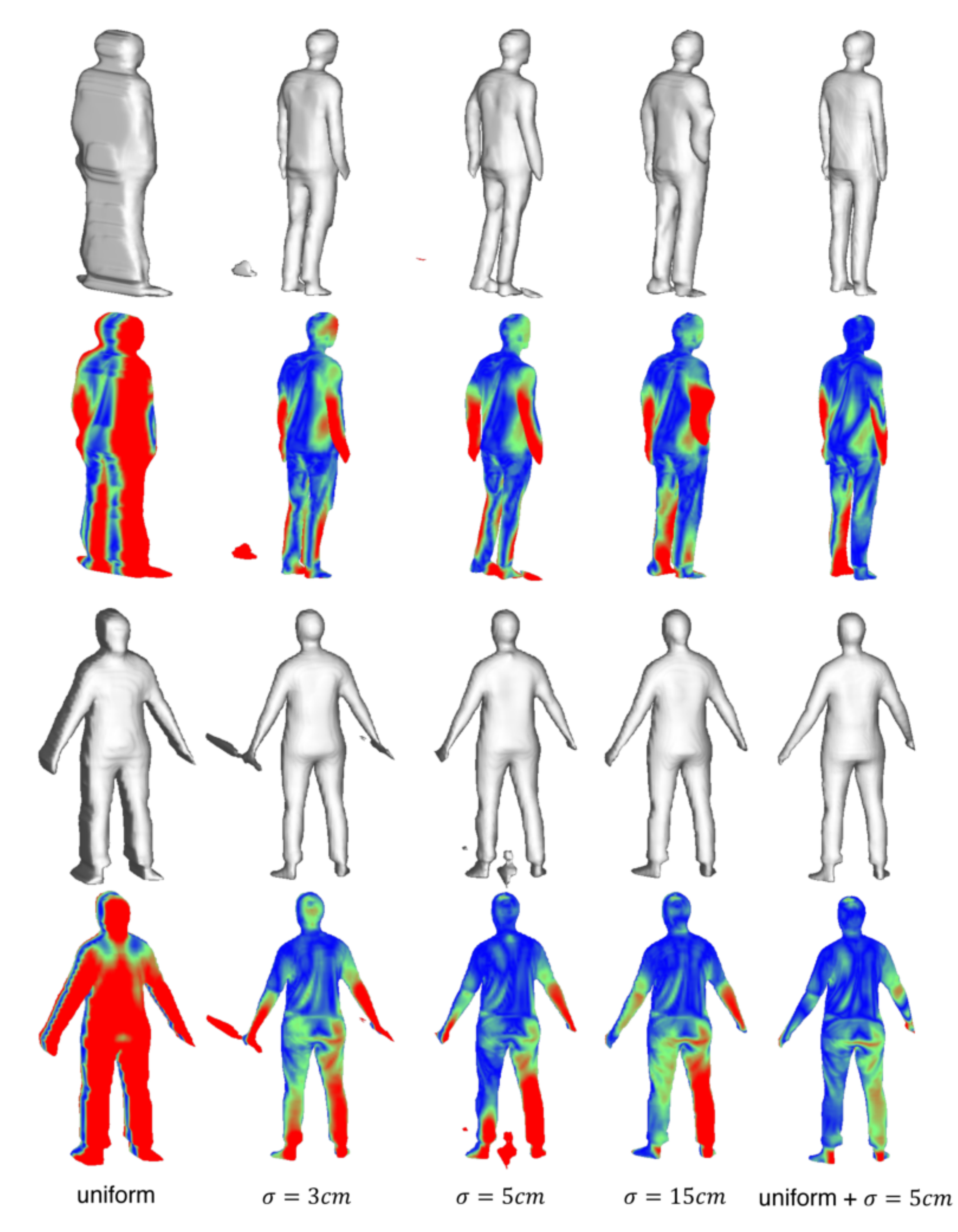}
\end{center}
\vspace{-10pt}
\caption{Reconstructed geometry and point to surface error visualization using different sampling methods.}
\label{fig:eval_sampling}
\end{figure}

\paragraph{Spatial Sampling.}
In Table \ref{tab:sampling} and Figure \ref{fig:eval_sampling}, we provide the effects of sampling methods for surface reconstruction. The most straightforward way is to uniformly sample inside the bounding box of the target object. Although it helps to remove artifacts caused by overfitting, the decision boundary becomes less sharp, losing all the local details (See Figure \ref{fig:eval_sampling}, first column). To obtain a sharper decision boundary, we propose to sample points around the surface with distances following a standard deviation $\sigma$ from the actual surface mesh. We use $\sigma=3, 5,$ and $15$ cm. The smaller $\sigma$ becomes, the sharper the decision boundary is the result becomes more prone to artifacts outside the decision boundary (second column). We found that combining adaptive sampling with $\sigma=5$ cm and uniform sampling achieves qualitatively and quantitatively the best results (right-most column). Note that each sampling scheme is trained with the identical setup as our training procedure described in Appendix I.

\begin{figure}[t]
\begin{center}
  \includegraphics[width=\linewidth]{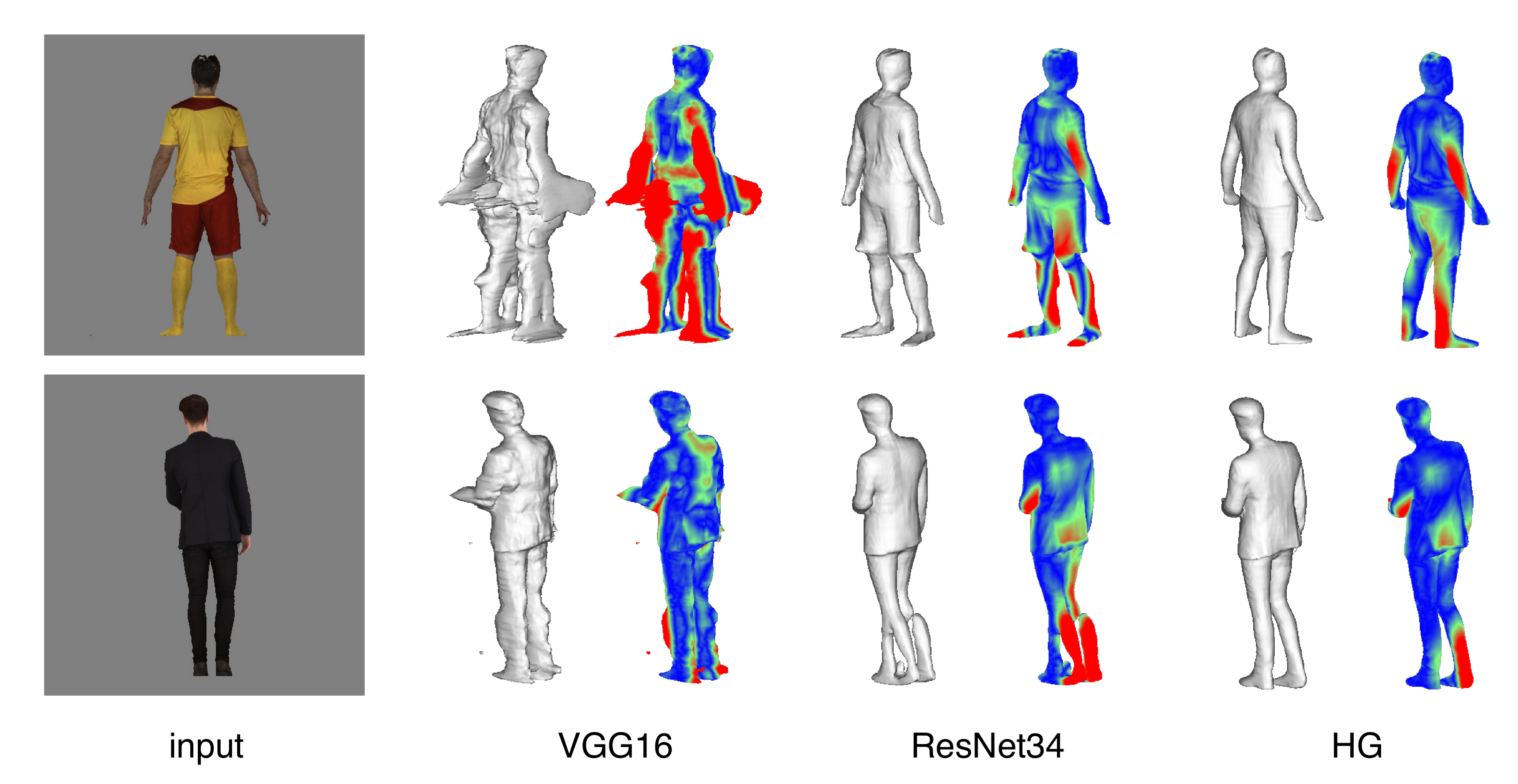}
\end{center}
\vspace{-10pt}
\caption{Reconstructed geometry and point to surface error visualization using different architectures for the image encoder.}
\label{fig:eval_architecture}
\end{figure}

\paragraph{Network Architecture.}
In this section, we show comparisons of different architectures for the surface reconstruction and provide insight on design choices of the image encoders. One option is to use bottleneck features of fully convolutional networks \cite{johnson2016perceptual, wei2016cpm, newell2016stacked}. Due to its state-of-the-art performance in volumetric regression for human faces and bodies, we choose Stacked Hourglass network \cite{newell2016stacked} with a modification proposed by \cite{jackson2018human}, denoted as HG. Another option is to aggregate features from multiple layers to obtain multi-scale feature embedding \cite{PixelNet, Huang18ECCV}. Here we use two widely used network architectures: VGG16 \cite{simonyan2014very} and ResNet34 \cite{he2016deep} for the comparison. We extract the features from the layers of `relu1\_2', `relu2\_2', `relu3\_3', `relu4\_3', and `relu5\_3' for VGG network using bilinear sampling based on $x$, resulting in $1472$ dimensional features. Similarly, we extract the features before every pooling layers in ResNet, resulting in $1024$-D features. We modify the first channel size in PIFu to incorporate the feature dimensions and train the surface reconstruction model using the Adam optimizer with a learning rate of $1\times10^{-3}$, the number of sampling of $10,000$ and batch size of $8$ and $4$ for VGG and ResNet respectively. Note that VGG and ResNet are initialized with models pretrained with ImageNet \cite{deng2009imagenet}. The other hyper-paremeters are the same as the ones used for our sequential network based on Stacked Hourglass.
    
In Table \ref{tab:architecture} and Figure \ref{fig:eval_architecture}, we show comparisons of three architectures using our evaluation data. While ResNet has slightly better performance in the same domain as the training data (i.e., test set in RenderPeople dataset), we observe that the network suffers from overfitting, failing to generalize to other domains (i.e., BUFF and DeepFashion dataset). Thus, we adopt a sequential architecture based on Stacked Hourglass network as our final model.

\begin{table}[t]
\resizebox{\columnwidth}{!}{
\begin{tabular}{llll|lll}
\hline
        & \multicolumn{3}{l|}{RenderPeople} & \multicolumn{3}{l}{Buff} \\
Methods & Normal     & P2S     & Chamfer    & Normal  & P2S         & Chamfer \\ \hline
Uniform & 0.119      & 5.07    & 4.23       & 0.132   & 5.98        & 4.53    \\
$\sigma=3$cm     & 0.104      & 2.03    & 1.62       & 0.114   & 6.15        & 3.81    \\
$\sigma=5$cm     & 0.105      & 1.73    & 1.55       & 0.115   & 1.54        & 1.41    \\
$\sigma=15$cm    & 0.100      & \textbf{1.49}    & \textbf{1.43}       & 0.105   & 1.37        & 1.26    \\
$\sigma=5$cm + \text{Uniform}    & \textbf{0.084}      & 1.52    & 1.50       & \textbf{0.092}   & \textbf{1.15}        & \textbf{1.14} \\ \hline
\end{tabular}
}
\caption{Ablation study on the sampling strategy.}
\label{tab:architecture}
\end{table}

\begin{table}[t]
\resizebox{\columnwidth}{!}{
\begin{tabular}{llll|lll}
\hline
        & \multicolumn{3}{l|}{RenderPeople} & \multicolumn{3}{l}{Buff} \\
Methods & Normal     & P2S     & Chamfer    & Normal  & P2S         & Chamfer \\ \hline
VGG16     & 0.125      & 3.02    & 2.25      & 0.144   & 4.65        & 3.08    \\
ResNet34    & 0.097      & \textbf{1.49}    & \textbf{1.43}       & 0.099   & 1.68        & 1.50    \\
HG    & \textbf{0.084}      & 1.52    & 1.50   & \textbf{0.092}   & \textbf{1.15}        & \textbf{1.14} \\ \hline
\end{tabular}
}
\caption{Ablation study on network architectures.}
\label{tab:sampling}
\end{table}

\section*{Appendix III. Additional Results}

Please see the supplementary video for more results.

\begin{figure}[t]
\begin{center}
  \includegraphics[width=0.9\linewidth]{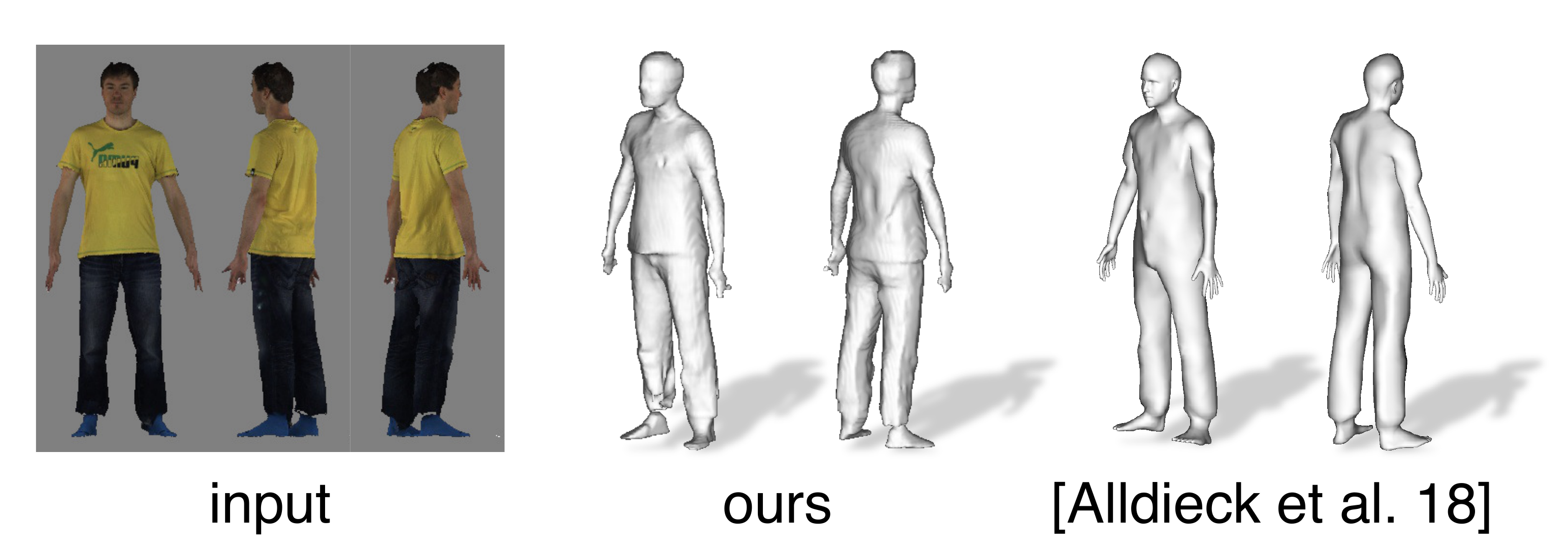}
\end{center}
\vspace{-15pt}
  \caption{
Comparison with a template-based method \cite{alldieck2018video}. Note that while Alldieck et al. uses a dense video sequence without camera calibration, ours uses the calibrated three views as input.
}
\vspace{-5pt}
\label{fig:videoavatar}
\end{figure}
\paragraph{Comparison with Template-based Method.}
\begin{table}[t]
\resizebox{\columnwidth}{!}{
\begin{tabular}{l|lll}
\hline
          &  \multicolumn{3}{c}{Buff} \\
Methods  & Normal    & P2S           & Chamfer \\ \hline
Alldieck et al. 18 (Video) &  0.127      & 0.820        & 0.795  \\
Ours (3 views)  & \textbf{0.107}      & \textbf{0.665}        & \textbf{0.641} \\
\hline
\end{tabular}
}
\caption{Quantitative comparison between a template-based method \cite{alldieck2018video} using a dense video sequence and ours using $3$ views.}
\label{tab:videoavatar}
\end{table}

In Figure \ref{fig:videoavatar} and Table \ref{tab:videoavatar}, we compare our approach with a template based method \cite{alldieck2018video} that takes a dense 360 degrees view video as an input on BUFF dataset. From 3 views we outperform the template based method. Note that Alldieck et al. requires an uncalibrated dense video sequence, while ours requires calibrated sparse view inputs.

\begin{figure}[t]
\begin{center}
  \includegraphics[width=0.95\linewidth]{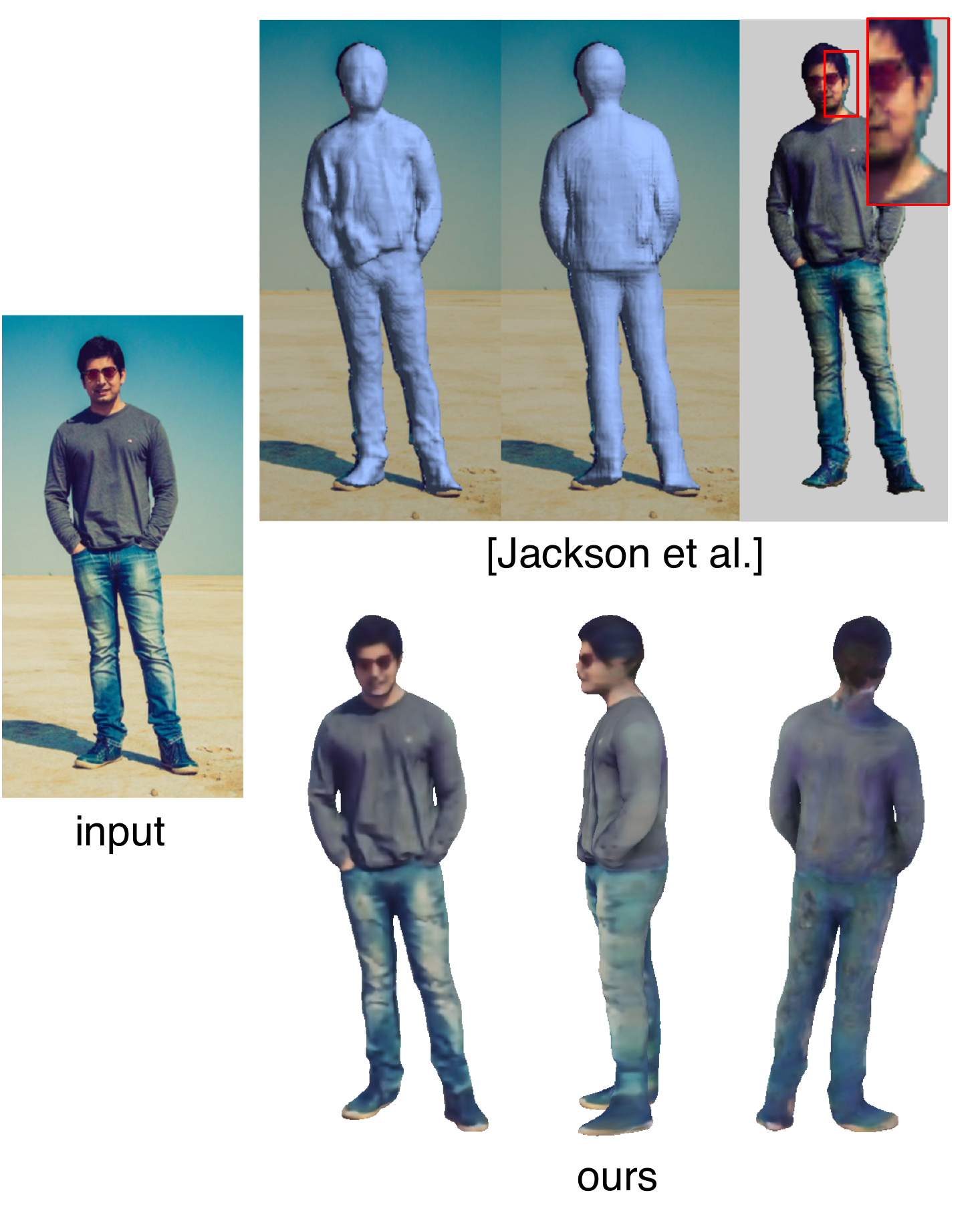}
\end{center}
\vspace{-10pt}
\caption{Comparison with Voxel Regression Network \cite{jackson2018human}. While \cite{jackson2018human} suffers from texture projection error due to the limited precision of voxel representation, our PIFu representation efficiently not only represents surface geometry in a pixel-aligned manner but also complete texture on the missing region. Note that \cite{jackson2018human} can only texture the visible portion of the person by projecting the foreground to the recovered surface. In comparison, we recover the texture of the entire surface, including the unseen regions. } 
\label{fig:comp_vrn}
\end{figure}

\paragraph{Comparison with Voxel Regression Network.}
We provide an additional comparison with Voxel Regression Network (VRN) \cite{jackson2018human} to clarify the advantages of PIFu. Figure \ref{fig:comp_vrn} demonstrates that the proposed PIFu representation can align the 3D reconstruction with pixels at higher resolution, while VRN suffers from misalignment due to the limited precision of its voxel representation. Additionally, the generality of PIFu offers texturing of shapes with arbitrary topology and self-occlusion, which has not been addressed by the work of VRN. Note that VRN only is able to project the image texture onto the recovered surface, and does not provide an approach to do texture inpainting on the unseen side. 

\begin{figure}[t]
\begin{center}
  \includegraphics[width=0.9\linewidth]{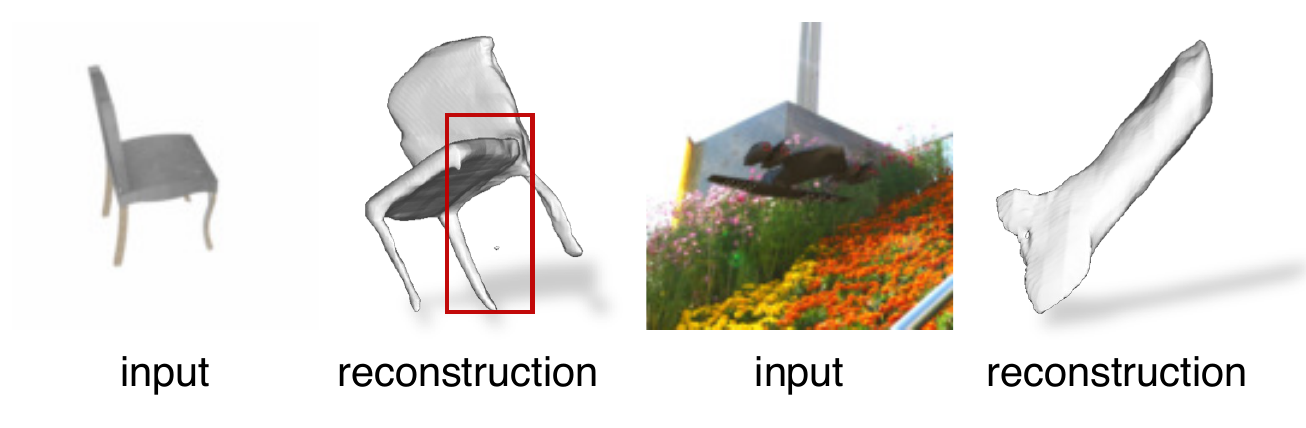}
\end{center}
\vspace{-15pt}
  \caption{
PIFu trained on general objects reveals new challenges to be addressed in future.
}
\vspace{-5pt}
\label{fig:generalobjects}
\end{figure}
\paragraph{Results on General Objects.}
In this work, we focused largely on clothed human surfaces. A natural question is how it extends to general object shapes. Our preliminary experiments on the ShapeNet dataset \cite{Chang:2015:ShapeNet} in a class agnostic setting reveals new challenges as shown in Figure~\ref{fig:generalobjects}. We speculate that the greater variety of object shapes makes it difficult to learn a globally coherent shape from only pixel-level features. Note that recently \cite{wang2019DISN} extend the idea of PIFu by explicitly combining global features and local features, demonstrating globally coherent and locally detailed reconstruction for general objects is possible. 

\begin{figure*}[t]
\begin{center}
  \includegraphics[width=0.9\textwidth]{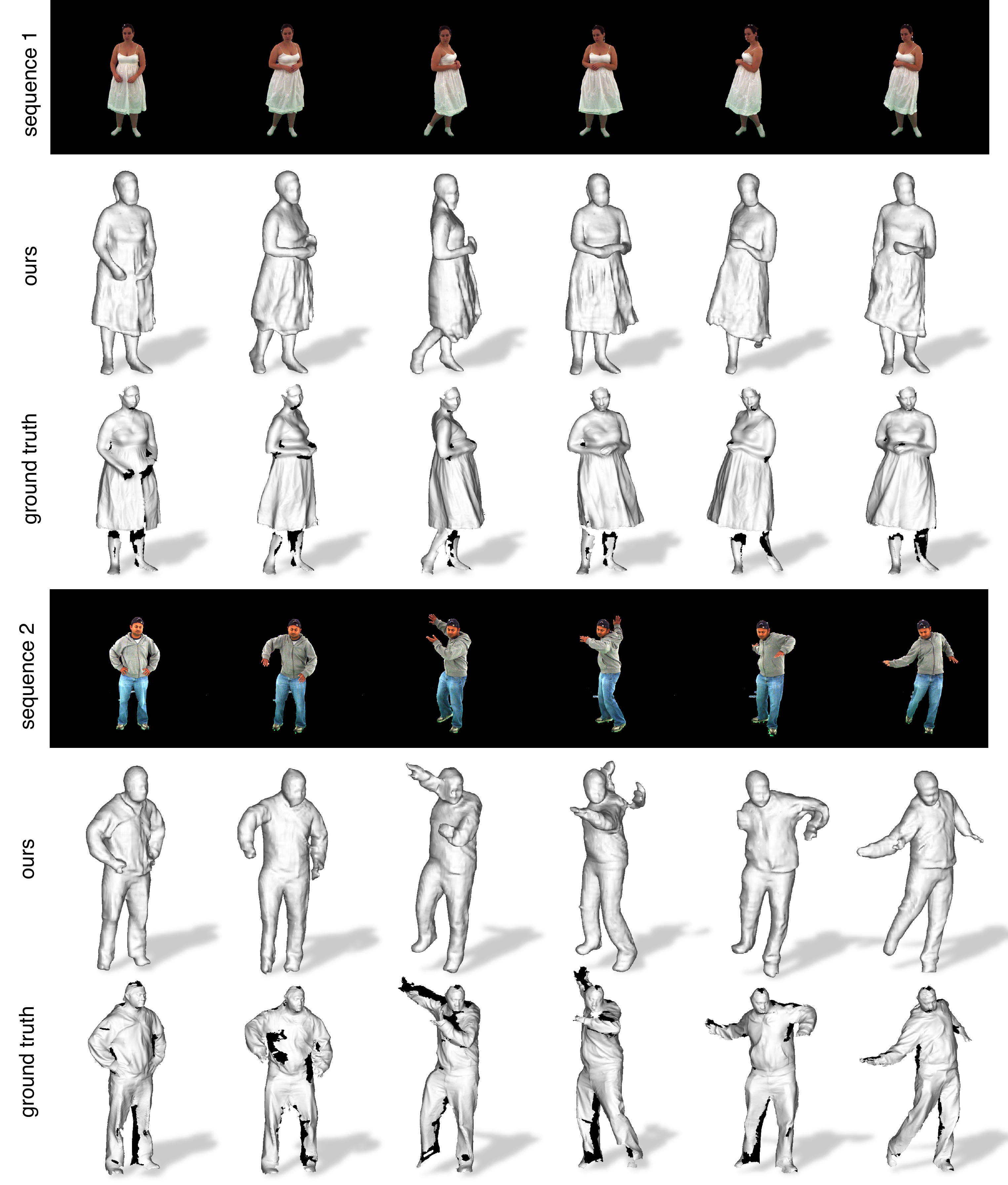}
\end{center}
\vspace{-10pt}
\caption{Results on video sequences obtained from \cite{vlasic2009dynamic}. While ours uses a single view input, the ground truth is obtained from $8$ views with controlled lighting conditions.}
\label{fig:video}
\end{figure*}

\paragraph{Results on Video Sequences.}
We also apply our approach to video sequences obtained from \cite{vlasic2009dynamic}. For the reconstruction, video frames are center cropped and scaled so that the size of the subjects are roughly aligned with our training data. Note that the cropping and scale is fixed for each sequence. Figure \ref{fig:video} demonstrates that our reconstructed results are reasonably temporally coherent even though the frames are processed independently.

\ifthenelse{\equal{\final}{0}}
{
}
{}
\end{document}